\pdfoutput=1

\documentclass[11pt]{article}

\usepackage{acl}

\usepackage{times}
\usepackage{latexsym}
\usepackage{booktabs}
\usepackage[autostyle]{csquotes}
\usepackage{graphicx}

\usepackage[T1]{fontenc}

\usepackage[utf8]{inputenc}

\usepackage{microtype}

\title{"You might think about slightly revising the title": Identifying Hedges in Peer-tutoring Interactions}

\author{Yann Raphalen$^{1}$, Chloé Clavel$^{2}$, Justine Cassell$^{1,3}$\\
$^{1}$Inria Paris \\  $^{2}$LTCI, Institut Polytechnique de Paris, Telecom-Paris \\ 
$^{3}$Carnegie Mellon University\\
\texttt{yann.raphalen.pro@gmail.com}, \texttt{justine@cs.cmu.edu}, \\ \texttt{chloe.clavel@telecom-paris.fr}}

\begin{document}
\maketitle
\begin{abstract}
Hedges play an important role in the management of conversational interaction. In peer-tutoring, they are notably used by tutors in dyads (pairs of interlocutors) experiencing low rapport to tone down the impact of instructions and negative feedback.
Pursuing the objective of building a tutoring agent that manages rapport with students in order to improve learning, we used a multimodal peer-tutoring dataset to construct a computational framework for identifying hedges. We compared approaches relying on pre-trained resources with others that integrate insights from the social science literature. Our best performance involved a hybrid approach that outperforms the existing baseline while being easier to interpret. We employ a model explainability tool to explore the features that characterize hedges in peer-tutoring conversations, and we identify some novel features, and the benefits of such a hybrid model approach. 
\end{abstract}

\section{Introduction}

Rapport, most simply defined as the “… relative harmony and smoothness of relations between people …” \cite{SpencerOatey2005}, has been shown to play a role in the success of activities as varied as psychotherapy \cite{Leach2005} and survey interviewing \cite{Lune2017}. In peer-tutoring, rapport, as measured by the annotation of thin slices of video, has been shown to be beneficial for learning outcomes \citep{Zhao2014towards, Sinha2015}. The level of rapport rises and falls with conversational strategies deployed by tutors and tutees at appropriate times, and as a function of the content of prior turns. These strategies include self-disclosure, referring to shared experience, and, on the part of tutors, giving instructions in an indirect manner. Some work has attempted to automatically detect these strategies in the service of intelligent tutors \cite{Zhao2016}, but only a few strategies have been attempted. Other work has concentrated on a "social reasoning module" \cite{Romero2017} to decide which strategies should be generated in a given context, but indirectness was not among the strategies targeted.  
\noindent In this paper, we focus on the automatic classification of one specific strategy that is particularly important for the tutoring domain, and therefore important for intelligent tutors: hedging, a sub-part of indirectness that "softens" what we say. This work is part of a larger research program with the long-term goal of automatically generating indirectness behaviors for a tutoring agent.
\begin{figure}[htbp]
    \centering
    \includegraphics[scale=0.2]{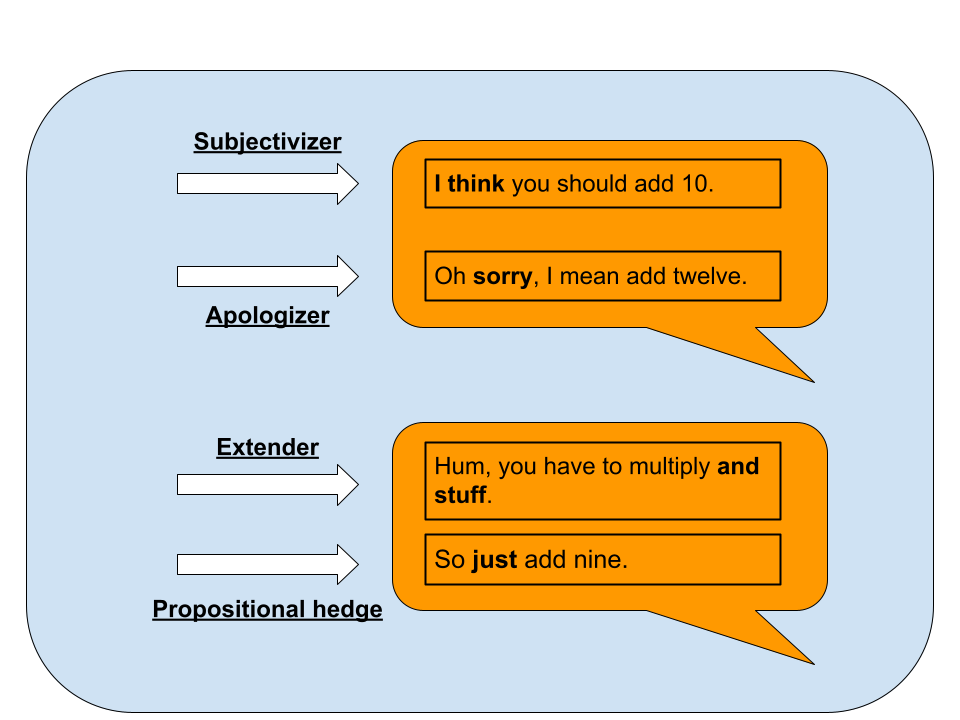}
    \caption{A mock conversation displaying each type of hedged formulation.}
    \label{fig:1}
\end{figure}

According to \citet{Brown1987}, hedges are part of the linguistic tools that interlocutors use to produce politeness, by limiting the face threat to the interlocutor (basically by limiting the extent to which the interlocutor might experience embarrassment because of some kind of poor performance). An example is "that's \textit{kind of} a wrong answer". Hedges are also found when speakers wish to avoid losing face themselves, for example when saying ("\textit{I think} I \textit{might} have to add 6."). \citet{Madaio2017} found that in a peer-tutoring task, when rapport between interlocutors is low, tutees attempted more problems and correctly solved more problems when their tutors hedged instructions, which likewise points towards a "mitigation of face threat" function.
\noindent Hedges can also be associated with a nonverbal component, for example averted eye gaze during criticism \cite{Burgoon1984}. Hedges are not, however, always appropriate, as in "I \textit{kind of think} it's raining today." when the interlocutors can both see rain (although it might be taken as humorous). These facts about hedges motivate a way to automatically detect them and, ultimately (although not in the current work) also generate them. In both cases we first have to be able to characterize them using interpretable linguistic features, which is what we address in the current paper. 
Thus, in the work described here, based on linguistic descriptions of hedges \citep{Brown1987, Fraser2010}, we built a rule-based classifier. We show that this classifier in combination with additional multimodal interpretable context-dependent features significantly improves the performance of a machine learning model for hedges, compared to a less interpretable deep learning baseline from \citet{Goel2019} using word embeddings. We also relied on a machine learning model explanation tool \cite{Lundberg2017} to investigate the linguistic features related to hedges in the context of peer-tutoring, primarily to see if we could discover surprising features that the classification model would associate to hedges in this context, and we describe those below. The code of the models described in the paper is also provided. \footnote{https://github.com/AnonymousHedges/HedgeDetection}

\section{Related work}

\noindent\textbf{Hedges:} 
\noindent According to \citet{Fraser2010}, hedging is a rhetorical strategy that attenuates the strength of a statement. One way to produce a hedge is by altering the full semantic value of a particular expression through \textbf{Propositional hedges} (also called \textbf{Approximators} in \citet{Prince1982}), as in "You are \textit{kind of} wrong," that reduce prototypicality (i.e accuracy of the correspondence between the proposition and the reality that the speaker seeks to describe). Propositional hedges are related to fuzzy language \cite{Lakoff1975}, and therefore to the production of vagueness \cite{williamson2002vagueness} and uncertainty \cite{Vincze2014}. \newline
A second kind are \textbf{Relational Hedges} (also called \textbf{Shields} in \citet{Prince1982}), such as “\textit{I think that} you are wrong.” or “\textit{The doctor wants you} to stop smoking.”, conveying that the proposition is considered by the speaker as subjective. In a further sub-division, \textbf{Attribution Shields}, as in "The doctor \textit{wants you ...}", the involvement of the speaker in the truth value of the proposition is not made explicit, which allows speakers not to take a stance. \newline
\noindent As described above, \citet{Madaio2017} found that tutors who showed lower rapport with their tutees used more hedged instructions (they also employed more positive feedback), however this was only the case for tutors with a greater belief in their ability to tutor. Tutees in this context  solved more problems correctly when their tutors hedged instructions. No effect of hedging was found for dyads (pairs of interlocutors) with greater social closeness. However, the authors did not look at the specific linguistic forms these teenagers used. \newline
\noindent \citet{Rowland2007} also describes the role that hedging plays in this age group, showing that students use both relational ("\textit{I think that} John is smart.") and propositional ("John is \textit{kind of} smart.") hedges for much the same shielding function of demonstrating uncertainty, to save them from the risk of embarrassment if they are wrong. The author observed that teens used few \textbf{Adaptors} (\textit{kind of}, \textit{somewhat}) and preferred to use \textbf{Rounders} (\textit{around}, \textit{close to}). However, this study was performed with an adult and two children, possibly biasing the results due to the participation of the adult investigator. Hedges have been included in virtual tutoring agents before now. \cite{Howard2015} integrated hedges in a tutor agent for undergraduates in CS, as a way to encourage the student to take the initiative. Hedges have also been used as a way of integrating Brown and Levinson's politeness framework \citep{Wang2008, Schneider2015} in virtual tutoring agents. Results were not broken out by strategy, but politeness in general was shown to positively influence motivation and learning, in certain conditions.\\ 
\noindent\textbf{Computational methods for hedge detection:} 
A number of studies have targeted the detection of hedges and uncertainty in text \citep{Medlock2007, Ganter2009, Tang2010, Velldal2011, Szarvas2012}, particularly following the CoNLL 2010 dataset release \cite{Farkas2010}. However, this work is not as related to hedges in conversation, as it focuses on a formal and academic language register \citep{Hyland1998, Varttala1999}. As noted by \citet{Prokofieva2014}, the functions of hedges are domain- and genre-dependent, therefore this bias towards formality implies that the existing work may not adapt well to the detection of hedges in conversation between teenagers. A consequence is that the existing work does not consider terms like "I think," since opinions rarely appear in an academic writing dataset. Instructions are also almost absent ("I think you have to add ten to both sides."), a strong limitation for the study of conversational hedges since it is in requests (including tutoring instructions) that indirect formulations mostly occur according to \citet{Blum1987}. 
\citet{Prokofieva2014} also note that it is difficult to detect hedges because the word patterns associated with them have other semantic and pragmatic functions: considering "I think that you have to add x to both sides." vs "I think that you are an idiot.", it is not clear that the second use of "I think that" is an hedge marker. They advocate using machine learning approaches to deal with the ambiguity of these markers.
Working on a conversational dataset, \citet{Ulinski2018} built a computational system to assess speaker commitment (i.e. at which point the speaker seems convinced by the truth value of a statement), in particular by relying on a rule-based detection system for hedges.
\noindent Compared to that work, our rule-based classification model is directly detecting hedge classes, and we employ the predictions of the rule-based model as a feature for stronger machine learning models, designed to lessen the impact of the imbalance between classes. We also consider \textbf{apologies} when they serve a mitigation function (we then call them \textbf{Apologizers}), as was done by the authors of our corpus, and we also use the term \textbf{subjectivizers} as defined below, to be able to compare directly with the previous work carried out on this corpus. 
As far as we know, only \citet{Goel2019} have worked with a peer-tutoring dataset (the same one that we also use), and they achieved their best classification result by employing an Attention-CNN model, inspired by \citet{Adel2017}. 

\section{Problem statement}

\noindent We consider a set D of conversations $D = (c_{1}, c_{2}, ..., c_{|D|})$, where each conversation is composed of a sequence of independent syntactic clauses $c_{i}= (u_{1},u_{2},...,u_{M})$, where M is the number of clauses in the conversation. Note that two consecutive clauses can be produced by the same speaker. 
Each clause is associated with a unique label corresponding to the different hedge classes described in Table~\ref{tab:2}: $y_{i} \in C$ = \{\textbf{Propositional Hedges}, \textbf{Apologizers}, \textbf{Subjectivizers}, \textbf{Not hedged}\}. Finally, an utterance $u_{i}$ can be represented as a vector of features $X = (x_{1}, x_{2}, ..., x_{N})$, where N represents the number of features we used to describe a clause. 
Our first goal is to design a model that correctly predicts the label $y_{i}$ associated to $u_{i}$. It can be understood as the following research question: 

\noindent\textbf{RQ1:} "Which models and features can be used to automatically characterize hedges in a peer-tutoring interaction?" 

\noindent Our second goal is to identify, for each hedge class, the set of features ${F_{class} = \{f_{k}\}}$, $k \in [1,N] $ sorted by feature importance in the classification of $class$. It corresponds to the following research question: 

\noindent\textbf{RQ2:} "What are the most important linguistic features that characterize our hedge classes in a peer-tutoring setting?" \newline

\section{Methodology}

\subsection{Corpus}
\noindent\textbf{Data collection:}
The dialogue corpus used here was collected as part of a larger study on the effects of rapport-building on reciprocal peer tutoring. 24 American teenagers (mean age = 13.5, min = 12, max = 15), half male and half female, came to a lab where half of the participants were paired with a same-age, same-gender friend, and the other half paired with a stranger. The participants were assigned to a total of 12 dyads in which the participants alternated tutoring one another in linear algebra equation solving for 5 weekly hour-long sessions, for a total corpus of nearly 60 hours of face-to-face interactions. Each session was structured such that the students engaged in brief social chitchat in the beginning, then one of the students was randomly assigned to tutor the other for 20 minutes. They then engaged in another social period, and concluded with a second tutoring period where the other student was assigned the role of tutor. Audio and video data were recorded, transcribed, and segmented for clause-level dialogue annotation, providing nearly 24 000 clauses. Non-speech segments (notably fillers and laughter) were maintained. Because of temporal misalignment for parts of the corpus, many paraverbal phenomena, such as prosody, were unfortunately not available to us.
\noindent Since our access to the dataset is covered by a Non-Disclosure Agreement, it cannot be released publicly. However the original experimenters' Institutional Review Board (IRB) approval allows us to view, annotate, and use the data to train models. This also allows us to provide a link to a pixelated video example in the GitHub repository of the project\footnote{https://github.com/AnonymousHedges/HedgeDetection}. \\
\noindent\textbf{Data annotation:}
The dataset was previously annotated by \citet{Madaio2017}, following an annotation manual that used hedge classes derived from \citet{Rowland2007} (see Table~\ref{tab:2}). Only the task periods of the interactions were annotated.  Comparing the annotations with the classes mentioned in the related work section, \textbf{Subjectivizers} correspond to \textbf{Relational hedges} \cite{Fraser2010}, \textbf{Propositional hedges} and \textbf{Extenders} correspond to \textbf{Approximators} \cite{Prince1982} with the addition of some discourse markers such as \textit{just}. \textbf{Apologizers} are mentioned as linguistic tools related to negative politeness in \citet{Brown1987}. 
\noindent Krippendorff’s alpha obtained for this corpus annotated by four coders was over 0.7 for all classes (denoting an acceptable inter-coder reliability according to \citet{Krippendorff2004}). 
The dataset is widely imbalanced, with more than 90\% of the utterances belonging to the \textbf{Not hedged} class. 

\noindent In reviewing the corpus and the annotation manual, however, we noticed two issues. First, the annotation of the  \textbf{Extenders} class was inconsistent, leading to the \textbf{Extenders} and \textbf{Propositional hedges} classes carrying similar semantic functions.  We therefore merged the two classes and grouped utterances labeled as \textbf{Extenders} and those labeled as \textbf{Propositional hedges} under the heading of \textbf{Propositional hedges}. 
Second, the annotation of clauses containing the tokens "just" and "would" (two terms occurring frequently in the dataset that are key components of \textbf{Propositional Hedges} and \textbf{Subjectivizers} but that are not in fact hedges in all cases) was also inconsistent, leading to virtually all clauses with those two tokens being considered hedges. We therefore re-considered all the clauses associated with any of the hedge classes, as well as all the clauses in the "Not hedged" class that contained "just" or "would". The re-annotation was carried out by two annotators who achieved a Krippendorff's alpha inter-rater  reliability of .9 or better for \textbf{Apologizers}, \textbf{Subjectivizers}, and \textbf{Propositional hedges} before independently re-annotating the relevant clauses. An example of a re-annotation was removing "I \textit{would} kill you!" from the hedge classes.

\begin{table*}
  \scriptsize
  \centering
  \setlength{\tabcolsep}{2pt}
  \begin{tabular}{@{}ccc@{}} \toprule
    Class & Definition & Example\\ \midrule
    Subjectivizers & Words that reduce intensity or certainty & “So then I would divide by two.” \\ 
    Apologizers & Apologies used to soften direct speech acts & “Oh sorry six b.” \\
    Propositional hedges & Qualifying words to reduce intensity or certainty of utterances & “It's actually eight.” \\
    Extenders & Words used to indicate uncertainty by referring to vague categories & “It'll be the number x or whatever variable you have.” \\ \bottomrule
  \end{tabular}
  \caption{Definition of the classes}
  \label{tab:2}
\end{table*}

\noindent 

\begin{table}[htbp]
  \small
  \centering
  \setlength{\tabcolsep}{2pt}
  \begin{tabular}{@{}ccccc@{}} \toprule
    Prop. hedges & Apologizers & Subjectivizers & Not hedged & Total\\ \midrule
    1210 & 128 & 626 & 21192 & 23156 \\ \midrule
  \end{tabular}
  \caption{Distribution of the classes}
  \label{tab:1}
\end{table}

\subsection{Features}

\noindent\textbf{Label from rule-based classifier (Label RB):} We use the class label predicted by the rule-based classifier described in Section~\ref{ssec:CM} as a feature. Our hypothesis is that the machine learning model can use this information to counterbalance the class imbalance. To take into account the fact that some rules are more efficient than others, we weighted the class label resulting from the rule-based model by the precision of the rule that generated it. 

\noindent\textbf{Unigram and bigram:} We count the number of occurrences of unigrams and bigrams of the corpus in each clause. We used the lemma of the words for unigrams and bigrams using the nltk lemmatizer (Loper, 2002) and selected unigrams and bigrams that occurred in the training dataset at least fifty times. The goal was to investigate, with a bottom-up approach, to what extent the use of certain words characterizes hedge classes in tutoring. In Section 5 we examine the overlap between these words and those \textit{a priori} identified by the rules.  

\noindent\textbf{Part-of-speech (POS):} Hedge classes seem to be associated with different syntactic patterns: for example, subjectivizers most often contain a personal pronoun followed by a verb, as in "I guess", "I believe", "I think". We therefore considered the number of occurrences of POS-Tag n-grams (n=1, 2, 3) as features. We used the spaCy POS-tagger and considered POS unigrams, bigrams and trigrams that occur at least 10 times in the training dataset.

\noindent\textbf{LIWC:} Linguistic Inquiry and Word Count (LIWC) \cite{Pennebaker2015} is standard software for extracting the count of words belonging to specific psycho-social categories (\textit{e.g.}, emotions, religion). It has been successfully used in the detection of conversational strategies \cite{Zhao2016}. We therefore count the number of occurrences of all the 73 categories from LIWC.

\noindent\textbf{Tutoring moves (TM):} Intelligent tutoring systems rely on specific tutoring moves to successfully convey content (as do human tutors). We therefore looked at the link between the tutoring moves, as annotated in \citet{Madaio2017}, and hedges. For tutors, these moves are (1) instructional directives and suggestions, (2) feedback, and (3) affirmations, mostly explicit reflections on their partners’comprehension, while for tutees, they are (1) questions, (2) feedbacks, and (3) affirmations, mostly tentative answers.

\noindent\textbf{Nonverbal and paraverbal behaviors:} As in \citet{Goel2019}, we included the nonverbal and paraverbal behaviors that are related to hedges. Specifically, we consider laughter and smiles, that have been shown to be effective methods of mitigation \cite{Warner2014}, cut-offs indicating self-repairs, fillers like "Um", gaze shifts (annotated as 'Gaze at Partner', 'Gaze at the Math Worksheet', and 'Gaze elsewhere'), and head nods. Each feature was present twice in the feature vector, one time for each interlocutor. 
\noindent Inter-rater reliability for nonverbal behavior was 0.89 (as measured by Krippendorff’s alpha) for eye gaze, 0.75 for smile count, 0.64 for smile duration and 0.99 for head nod. Laughter is also reported in the transcript at the word level. We separate the tutor's behaviors from those of the tutee. The collection process for these behaviors is detailed further in \citet{Zhao2016socially}.

\begin{table}[t]
  \tabcolsep=0.15cm
  \small
  \centering
  \begin{tabular}{ccc} \toprule
    Features name & Automatic extraction & Vector size \\ \midrule
    Rule-based label & Yes & 4 \\
    Unigram & Yes & \textasciitilde 250 \\
    Bigram & Yes & \textasciitilde 250\\
    POS & Yes & \textasciitilde 1200 \\
    LIWC & Yes & 73 \\
    Nonverbal & No & 24 \\
    Tutoring moves & No & 6 \\ 
    Total & & \textasciitilde 1800 \\\bottomrule
  \end{tabular}
  \caption{List of automatically extracted and manually annotated features with their size.}
  \label{tab:3}
\end{table}

\noindent The clause-level feature vector was normalized by the length of the clause (except for the rule-based label). This length was also added as a feature. Table~\ref{tab:3} presents an overview of the final feature vector. 
 
\subsection{Classification models}
\label{ssec:CM}
The classification models used are presented here according to their level of integration of external linguistic knowledge.

\noindent\textbf{Rule-based model:} On the basis of the annotation manual used to construct the dataset from \citet{Madaio2017}, and with descriptions of hedges from \citet{Rowland2007}, \citet{Fraser2010} and \citet{Brown1987}, we constructed a rule-based classifier that matches regular expressions indicative of hedges. The rules are detailed in Table~\ref{tab:7} in the Appendix. 

\noindent\textbf{LGBM:} Since hedges are often characterized by explicit lexical markers, we tested the assumption that a machine learning model with a knowledge-driven representation for clauses could compete with a BERT model in performance, while being much more interpretable. We relied on LightGBM, an ensemble of decision trees trained with gradient boosting \cite{Ke2017}. This model was selected because of its performance with small training datasets and because it can ignore uninformative features, but also for its training speed compared to alternative implementations of gradient boosting methods.

\noindent\textbf{Multi-layer perceptron (MLP):} As a simple baseline, we built a multi-layer perceptron using three sets of features: a pre-trained contextual representation of the clause (SentBERT; \citet{Reimers2019}) ; the concatenation of this contextual representation of the clause and a rule-based label (not relying on the previous clauses) ; and finally the concatenation of all the features mentioned in section 4.2, without the contextualized representation.

\noindent\textbf{LSTM over a sequence of clauses:} Since we are working with conversational data, we also wanted to test whether taking into account the previous clauses helps to detect the type of hedge class in the next clause.
\noindent Formally, we want to infer $y_{i}$ using $y_{i} = \max_{y \in Classes}P(y|X(u_{i}), X(u_{i-1}), ..., X(u_{i-K}))$ , where K is the number of previous clauses that the model will take into account. The MLP model presented above infers $y_{i}$ using $y_{i} = \max_{y \in Classes}P(y|X(u_{i}))$, therefore a difference of performance between the two models would be a sign that using information from the previous clauses could help to detect the hedged formulation in the current clause. 
\noindent We tested a LSTM model with the same representations for clauses as for the MLP model.

\noindent\textbf{CNN with attention:} \citet{Goel2019} established their best performance on hedge detection using a CNN model with additive attention over word (and not clause) embeddings. 
Contrary to the MLP and LSTM models mentioned above, this model tries to infer $y_{i}$ using $y_{i} = \max_{y \in Classes}P(y|g(w_{0}), g(w_{1}), ..., g(w_{L}))$, with L representing the maximum clause length we allow, and g representing a function that turns the word $w_{j},\ j \in [0, L]$ into a vector representation (for more details, please see \citet{Adel2017}).

\noindent\textbf{BERT:} To benefit from deep semantic and contextual representations of the utterances, we also fine-tuned BERT \cite{Devlin2019} on our classification task. BERT is a pre-trained Transformers encoder \cite{Vaswani2017} that has significantly improved the state of the art on a number of NLP tasks, including sentiment analysis. It produces a contextual representation of each word in a sentence, making it capable of disambiguating the meaning of words like "think" or "just" that are representative of certain classes of hedges. BERT, however, is notably hard to interpret.

\subsection{Analysis tools}

Looking at which features improve the performance of our classification models tells us whether these features are informative or not, but does not explain how these features are used by the models to make a given prediction. We therefore produced a complementary analysis using an interpretability tool. As demonstrated by \cite{Lundberg2017}, LightGBM internal feature importance scores are inconsistent with both the model behavior and human intuition, so we instead used a model-agnostic tool. 
\noindent SHAP \cite{Lundberg2017} assigns to each feature an importance value (called Shapley values) for a particular prediction depending on the extent of its contribution (a detailed introduction to Shapley values and SHAP can be found in \citet{Molnar2020}). SHAP is a model-agnostic framework, therefore the values associated with a set of features can be compared across models. It should be noted that SHAP produces explanations on a case-by-case basis, therefore it can both provide local and global explanations. For the Gradient Boosting model, we use an adapted version of SHAP \cite{Lundberg2018}, called TreeSHAP.

\section{Experiments and results}

\subsection{Experimental setting}

\noindent To detect the best set of features, we used LightGBM and proceeded incrementally, by adding the group of features we thought to be most likely associated with hedges. We did not consider the risk of relying on a sub-optimal set of features through this procedure because of the strong ability of LightGBM to ignore uninformative features. 
We use this incremental approach as a way to test our intuition about the performativity of groups of features (\textit{i.e.} does adding a feature improve the performance of the model) with regard to the task of classification. To compare our models, we trained them on the 4-class task, and looked at the average of the weighted F1-scores for the three hedge classes (\textit{i.e.} how well the models infer minority classes) that we report here as "3-classes", and at the average of the weighted F1-scores for the 4 classes, that we report as "4-classes". Details of the hyperparameters and experimental settings are provided in Appendix \ref{appendix:Parameters}.

\subsection{Model comparison and feature analysis}
\noindent\textbf{Overall results:} Table~\ref{tab:4} presents the results obtained by the 6 models presented in Section~\ref{ssec:CM} for the multi-class problem. 
Best performance (F1-score of 79.0) is obtained with LightGBM leveraging almost all the features. In the appendix (see Table~\ref{tab:8} and Table~\ref{tab:9}) we indicate the confidence intervals to represent the significance of the differences between the models. 

\noindent First, and perhaps surprisingly, we notice that the use of "Knowledge-Driven" features based on rules built from linguistic knowledge of hedges in the LightGBM model outperforms the use of pre-trained embeddings within a fine-tuned BERT model (79.0 vs. 70.6), and in the neural baseline from \cite{Goel2019} (79.0 vs 64.5).

\noindent The low scores obtained by the LGBM, LSTM and MLP models with pre-trained sentence embeddings versus Knowledge-Driven features might signal that the word patterns characterizing hedges are not salient in these representations (i.e. the distance between "\textit{I think} you should add 5." and "You should add 5." is short.). KD Features seem to provide a better separability of the classes. The combination of KD features and Pre-trained embeddings does not significantly improve the performance of the models compared to the KD Features only, which suggests that the information from the Pre-trained embeddings is redundant with the one from the KD Features. This result may be due to the high dimensionality of the input vector (868 with PCA on the KD Features; ~ 2500 otherwise).

\noindent A second finding is that the use of gradient boosting models on top of rule-based classifiers better models the hedge classes. The other machine learning models did not prove to be as effective, except for BERT.

\begin{table}
  \tabcolsep=0.12cm
  \tiny
  \centering
  \begin{tabular}{c|ccc} \toprule 
    Models & KD Feat. (KDF) & Pre-Trained Emb. (PTE) & KDF + PTE \\ \midrule
    Rule-based (3-classes) &  67.6 & $\emptyset$ & $\emptyset$ \\
    MLP (3-classes) &  68.5 (1.6) & 35.8 (3.1) & 64.8 (1.1) \\
    Attention-CNN (3-classes) & $\emptyset$ &  64.5 (3.0) & $\emptyset$ \\
    LSTM (3-classes) &  65.1 (5.7) &  39.8 (8.0) &  65.2 (5.1) \\
    BERT (3-classes) & $\emptyset$ &  \textbf{70.6 (2.3)} & $\emptyset$ \\
    LGBM (3-classes) &  \textbf{79.0 (1.3)}  &  35.0 (2.2) &  \textbf{70.1 (1.4)} \\ \midrule
    Rule-based (4-classes) &  94.7 & $\emptyset$ & $\emptyset$ \\
    MLP (4-classes) &   94.8 (0.3) &  89.7 (0.4) &  93.9 (0.4) \\
    Attention-CNN (4-classes) & $\emptyset$ &  94.4 (0.2) & $\emptyset$ \\
    LSTM (4-classes) &  93.9 (1.4) &  89.1 (1.4) &  94.1 (1.2) \\
    BERT (4-classes) & $\emptyset$ &  \textbf{94.9 (0.4)} & $\emptyset$ \\
    LGBM (4-classes) &  \textbf{96.7 (0.2)}  &  91.0 (0.2) &  \textbf{95.4 (0.2)} \\ \bottomrule
  \end{tabular}
  \caption{Averaged weighted F1-scores (and standard deviation) for the three minority classes and for the 4 classes, for all models. "KD" stands for "Knowledge-Driven", meaning that the features are derived from lexicon, n-gram models and annotations.}
  \label{tab:4}
\end{table}

\noindent\textbf{Feature analysis using LightGBM:} Using the best performing model, Table~\ref{tab:5} shows the role of each feature set in the prediction task. The significance of the differences is shown in Table~\ref{tab:10} and Table~\ref{tab:11}. Compared to the rule-based model, the introduction of n-grams significantly improved the performance of our classifier, suggesting that some lexical and syntactic information describing the hedge classes was not present in the rule-based model. Looking at Table~\ref{tab:5}, we do not observe significant differences between the LGBM model using only the label rule based + (1-grams and 2-grams) and the models incorporating more features. To our surprise, neither the tutoring moves nor the nonverbal features significantly improved the performance of the model. The 2 features were included to index the specific peer tutoring context of these hedges, so this indicates that in future work we might wish to apply the current model to another context of use to see if this model of hedges is more generally applicable than we originally thought. By combining this result with the increased performance of the model using Knowledge-Driven (\textit{i.e.} explicit) features compared to pre-trained embeddings, it would seem that hedges are above all a lexical phenomenon (\textit{i.e.} produced by specific lexical elements).

\begin{table*}[t]
  \tabcolsep=0.15cm
  \footnotesize
  \centering
  \begin{tabular}{c|cccccccc} \toprule
    Models & Label RB & + 1-gram and 2-gram & + POS & + LIWC & + TM  & + Nonverbal\\ \midrule
    3-classes & 68.8 (0.8) & 78.2 (1.6) & 78.1 (1.3) & 79.0 (1.3) & 78.5 (2.4) & 78.7 (1.8) \\ 
    \midrule
    4-classes & 95.0 (0.2) & 96.5 (0.3) & 96.5 (0.2) & 96.7 (0.2) & 96.6 (0.4) & 96.7 (0.3)\\
    \bottomrule
  \end{tabular}
  \caption{Averaged weighted F1-scores for the three classes of hedges and the four classes, with an additive integration of KDF features in the LightGBM model. The standard deviation is computed across five folds.}
  \label{tab:5}
\end{table*}

\subsection{In-depth analysis of the informative features}

\noindent We trained the SHAP explanation models on LightGBM with all features. The most informative features (in absolute value) for each class are shown in Table 6, and the plots by class are presented in the Appendix. The most important features seem to be the rule-based labels, which appear in at least the fourth position for three classes (see Table~\ref{tab:6}), and in the first position for \textbf{Propositional Hedges} and \textbf{Not hedged} classes. Surprisingly, the Rule-Based label does not appear in the top 20 features for \textbf{Apologizers}. However, given that the class rarely appears in the data, the rules seldom activate, so the feature may simply be informative for a very small number of clauses. Unigrams (\textit{Oh}, \textit{Sorry}, \textit{just}, \textit{Would}, and \textit{I}) are also present in the 5 top-ranked features. This confirms the findings mentioned in related work for the characterization of the different hedge classes (\textit{just} with \textbf{Propositional Hedges}, \textit{sorry} with \textbf{Apologizer}, \textit{I} with \textbf{Subjectivizers}). The presence of \textit{Oh} also has high importance for the characterization of \textbf{Apologizer} (n=2), as illustrated in examples such as "\textit{Oh} sorry, that's nine.". We note that the occurrences of "\textit{Oh sorry}" as a stand-alone clause were excluded by our rule-based model because they do not correspond to an apologizer (they cannot mitigate the content of a proposition if there is no proposition associated). This example illustrates the interest of a machine learning model approach to disambiguate the function of conventional non-propositional phrases like "\textit{Oh sorry}".

\begin{table*}[t]
  \tabcolsep=0.08cm
  \footnotesize
  \centering
  \begin{tabular}{lcccccc} \toprule
    Rank & Apologizer & Subjectivizers & Prop. Hedges & Not hedged \\ \midrule
    1 & Function words (LIWC) & "I" & Class label & Class label \\
    2 & "Oh" (LIWC) & "Yeah" & "Would" & "Would" \\
    3 & "Sorry" & Noun (POS) & "Just" & "Yeah" \\
    4 & Affect (LIWC) & Class label & Function word (LIWC) & Noun (POS) \\
    5 & Clause length & Cognitive process (LIWC) & Netspeak (LIWC) & Cognitive process (LIWC) \\
  \end{tabular}
  \caption{Most important clause-level features for LightGBM according to the SHAP analysis.}
  \label{tab:6}
\end{table*}

\noindent In addition, SHAP highlights the importance of novel features whose function was not identified in the hedges literature: \textit{(i)} what LIWC classifies as \textbf{informal words} but that are mostly interjections like \textit{ah} and \textit{oh} are strongly associated with \textbf{Apologizer}, as are disfluencies (n=12); \textit{(ii)} the use of \textbf{POS tags} seems to be very relevant for characterizing the different classes (2-gram of POS tag features\footnote{Note that there is strong redundancy between some features of LIWC and the spaCy POS tagger that both produce a "Pronoun" category, using a lexicon in the first case, and a neural inference in the second.} occur in the top-ranked features of all the classes (see Figures in the Appendix). It means that there are some recurring syntactic patterns in each class; \textit{(iii)} Regarding the \textbf{utterance size}, a clause shorter than the mean is weakly associated with \textbf{directness} (n=17) while a longer clause suggests that it contains a \textbf{Subjectivizer (n=6)}. \textbf{Apologizers} are characterized by a mean clause length (n=5), with few variations from it; \textit{(iv)} Tutoring moves are not strong predictors of any classes: "Affirmation from tutor" is the only feature appearing as a predictor of Propositional hedges (n=20). This is consistent with the feature analysis in Table~\ref{tab:5}, suggesting that tutoring moves do not significantly improve the performance of the classifier; \textit{(v)} \textbf{Nonverbal behaviors} do not appear as important features for the classification. This is coherent with results from \cite{Goel2019}. Note that prosody might play a role in detecting instructions that trail off, but, as described, paraverbal features were not available; \textit{(vi)} \textit{Would} plays an important role in the production of hedges, as it is strongly associated to \textbf{Propositional hedges} (n=2). It is interesting to note that, when designing the rule-based classifier, we saw it decrease in performance when we started to include \textit{would} in our regular expression patterns, probably because the form is hard to disambiguate for a deterministic system.

\noindent While exploring the Shapley values associated to each clause, we observed that features like tutoring moves are extremely informative for a very small number of clauses (therefore not significantly influencing the overall performance of the prediction), and more or less not informative for the rest. Inferring the global importance of a feature as a mean across the shapley values in the dataset may not be the only way to explore the behavior of gradient boosting methods. It might be more useful to cluster clauses based on the importance that SHAP gives to that feature in its classification, as this could help discover sub-classes of hedges that are differentiated from the rest by their interaction with a specific feature (in the way that some \textbf{Apologizers} are characterized by an "oh").
\noindent We also note that the explanation model is sensitive to spurious correlations in the dataset, caused by the small representation of some class: for example, "nine" (n=7) and "four" (n=20) are positive predictors of \textbf{Apologizers}.

\section{Conclusion and future work}

\noindent Through our classification performance experiments, we showed that it is possible to use machine learning methods to diminish the ambiguity of hedges, and that the hybrid approach of using rule-based label features derived from social science (including linguistics) literature within a machine learning model helped significantly to increase the model's performance.  
Nonverbal behaviors and tutoring moves did not provide information at the sentence level; both the performance of the model and the feature contribution analysis suggested that their impact on the model output was not strong. This is consistent with results from \citet{Goel2019}. However, in future work we would like to investigate the potential of multimodal patterns when we are able to better model sequentiality (\textit{e.g.}, negative feedback followed by a smile).
\noindent Regarding the SHAP analysis, most of the features that are considered as important are coherent with the definition of the classes (\textit{I} for subjectivizers, \textit{sorry} for apologizers, \textit{just} for propositional hedges). However, we discovered that features like utterance size can also serve as indicators of certain classes of hedges.
\noindent A limitation of SHAP is that it makes a feature independence assumption, which prompts the explanatory model to underestimate the importance of redundant features (like pronouns in our work). In the future we will explore explanatory models capable of taking into account the correlation between features in the dataset like SAGE \cite{Covert2020}, but suited for very imbalanced datasets.
\noindent In the domain of peer-tutoring, we would like to be able to further test the link between hedges and rapport, and the link between hedges and learning gains in the subject being tutored. As noted above, this kind of study requires a fine-grained control of the language produced by one of the interlocutors, which is difficult to achieve in a human-human experience. 

\noindent We note that the hedge classifier can be used not just to classify, but also to work towards improving the generation of hedges for tutor agents. In future work we will explore using the classifier to re-rank generation outputs,  taking advantage of the recurring syntactic patterns (see \textit{(ii)} in Section 5.3) to improve the generation process of hedges, and re-generating clauses that don’t contain one of these syntactic patterns.

\section*{\textbf{Acknowledgments}}
Many thanks to members of the ArticuLabo at INRIA Paris for their precious assistance. This work was supported in part by the the French government under management of Agence Nationale de la Recherche as part of the “Investissements d’avenir” program, reference ANR-19-P3IA-0001 (PRAIRIE 3IA Institute). 

\bibliography{bibtext}
\bibliographystyle{acl_natbib}

\newpage

\appendix

\section{Additional information on the experimental settings}
\label{appendix:Parameters}

\noindent We used PyTorch \cite{Paszke2019} to implement the neural models. For each set of features, hyperparameters were selected using Optuna (Akiba, 2019), a parameter search framework. We re-implemented the Attention-CNN with Glove \cite{pennington2014glove} 300-D words embeddings as the vector representation. For each models, the results are cross-validated using 5 folds (we chose 5 instead of 10 to avoid having folds with too few samples per class). We corrected the loss function for class imbalance to force the model to adapt more to the less frequent classes. The strength of this correction depended on the model, and was selected because it provided a satisfying compromise between favoring recall and precision in the classification results of that model. For LightGBM, a "square root of the square root of the inverse class proportion" correction was selected. Neural models were trained using AdamW as an optimizer \cite{Loshchilov2018}, and used a reduced feature vector, obtained with the application of PCA ($d_{init}$ ~= 1800; d = 100 ; 99.8 {\%} of the information is conserved). No significant performance differences were observed between the original vector and the reduced vector for training the models. To compute the SHAP values mentioned in the paper, we kept one split to perform the 5-split of the dataset, and leave 1 split to validate and early stop the model, in order to avoid overfitting. A complete configuration of hyperparameters used for each model is reported in the GitHub repository with the code of the paper: https://github.com/YannRaphalen/Hedges-Detection.

\noindent The BERT model was fine-tuned on a Nvidia Quadro RTX 8000 GPU.

\section{Tables}

\begin{table*}[t]
  \tabcolsep=0.11cm
  \small
  \centering
  \begin{tabular}{l|c} \toprule
    Class & Rule (regexp) \\ \midrule
    Subj. & (?!what).*(i|we) ?(don't|didn't|did)? ?(not)? \\    & (guess|guessed|thought|think|believe|believed|suppose|supposed) \\ & ?(whether|if|is|that|it|this)?.* \\
    Subj. & .*(i|i'm|we) ?(was|am|wasn't)? ?(not)? (sure|certain).* \\
    Subj. & .*(i feel like you).* \\
    Subj. & .*(you (might|may) (believe|think)).* \\
    Subj. & .*(according to|presumably).* \\
    Subj. & .*(i|you|we) have to (check|look|verify).* \\
    Subj. & .*(if i'm not wrong|if i'm right|if that's true).* \\
    Subj. & .*(unless i).* \\
    Apol. & .*(i'm|i|we're) (am|are)? ?(apologize|sorry).* \\
    Apol. & (?!.*(be|been|was) like excuse me)((excuse me|sorry)[w ,']+|[w ,']+(excuse me|sorry)) \\
    Prop. & .*(just|a little|maybe|actually|sort of|kind of|pretty \\ & much|somewhat|exactly|almost|little bit|quite| \\ &
    regular|regularly|actually|almost|as it were|basically| \\ &
    probably|can be view as|crypto-|especially|essentially| \\ &
    exceptionally|for the most part|in a manner of speaking| \\ &
    in a real sense|in a sense|in a way|largely|literally| \\ &
    loosely speaking|kinda|more or less|mostly|often| \\ &
    on the tall side|par excellence|particularly| \\ &
    pretty much|principally|pseudo-|quintessentially| \\ &
    relatively|roughly|so to say|strictly speaking| \\ &
    technically|typically|virtually|approximately| \\ &
    something between|essentially|only).* \\
    Prop. & .*(i|i'm|you|it's) (am|are) (apparently|surely)[ ,]?.* \\ 
    Prop. & .*(it) (looks|seems|appears)[ ,]?.*", ".* (or|and)  (that|something|stuff|so forth) \\
  \end{tabular}
  \caption{Regexp rules used for the classifier.}
  \label{tab:7}
\end{table*}

\begin{figure*}[t]
    \centering
    \includegraphics[scale=0.5]{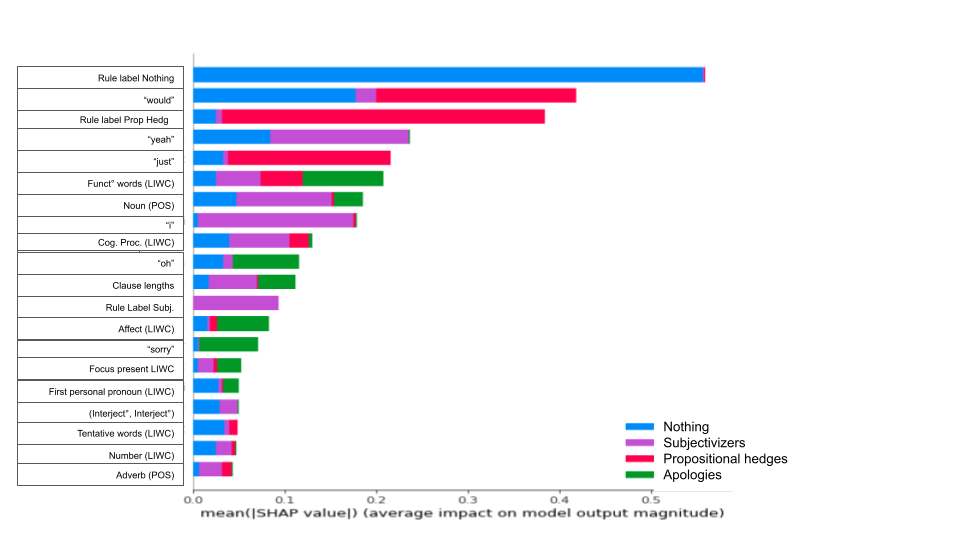}
    \caption{Absolute averaged feature contribution, as indicated by SHAP. The longer the bar is for one color, the more the feature is associated with the class represented by that color.}
    \label{fig:2}
\end{figure*}

\begin{figure*}[t]
    \centering
    \includegraphics[scale=0.5]{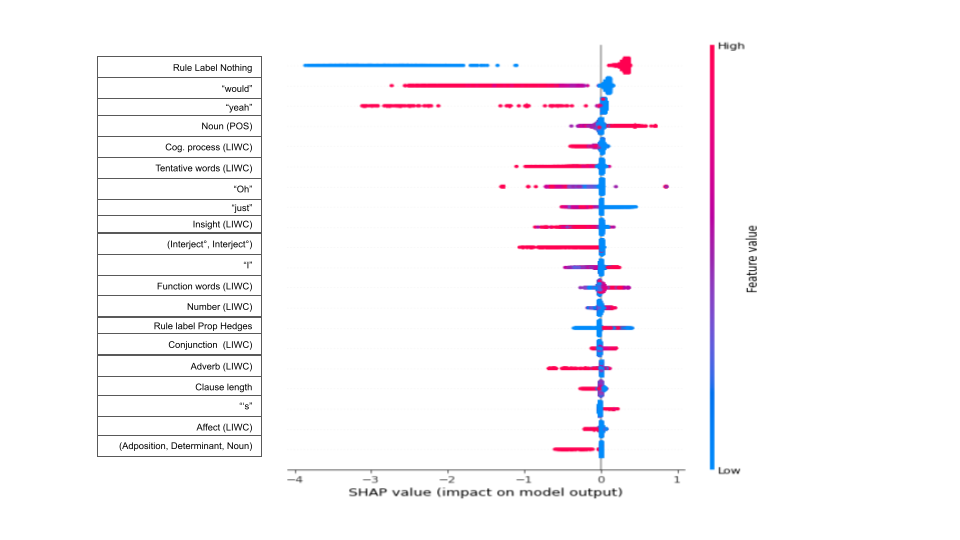}
    \caption{Averaged contribution of features to the detection of the "Not indirect" class, as indicated by SHAP. Each dot corresponds to a classified clause. A red dot indicates that the feature is present in the clause, while a blue dot indicates that the feature is absent. The farther on the right the dot is, the more the feature contributed to its classification as a hedge.}
    \label{fig:3}
\end{figure*}

\begin{figure*}[t]
    \centering
    \includegraphics[scale=0.5]{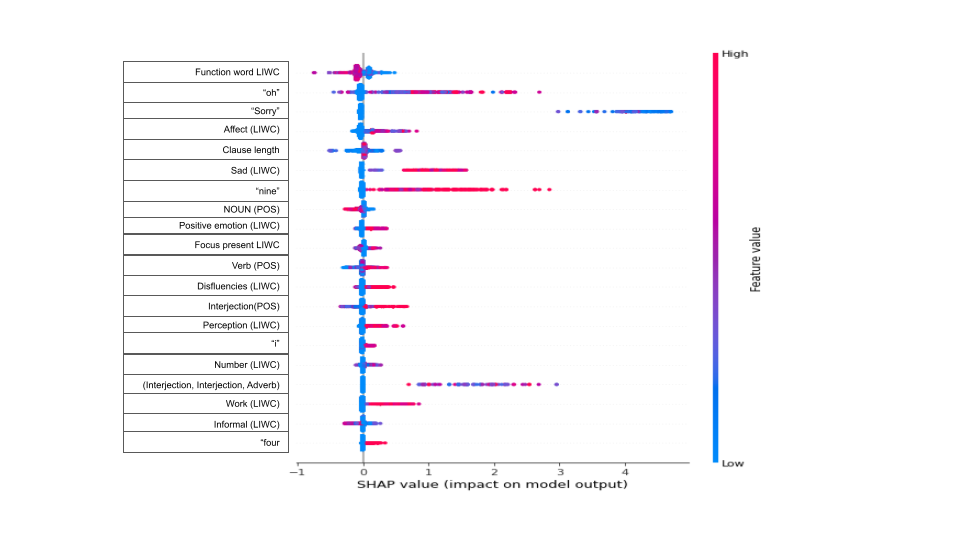}
    \caption{Averaged contribution of features to the detection of "Apologizers", as indicated by SHAP.}
    \label{fig:4}
\end{figure*}

\begin{figure*}[t]
    \centering
    \includegraphics[scale=0.5]{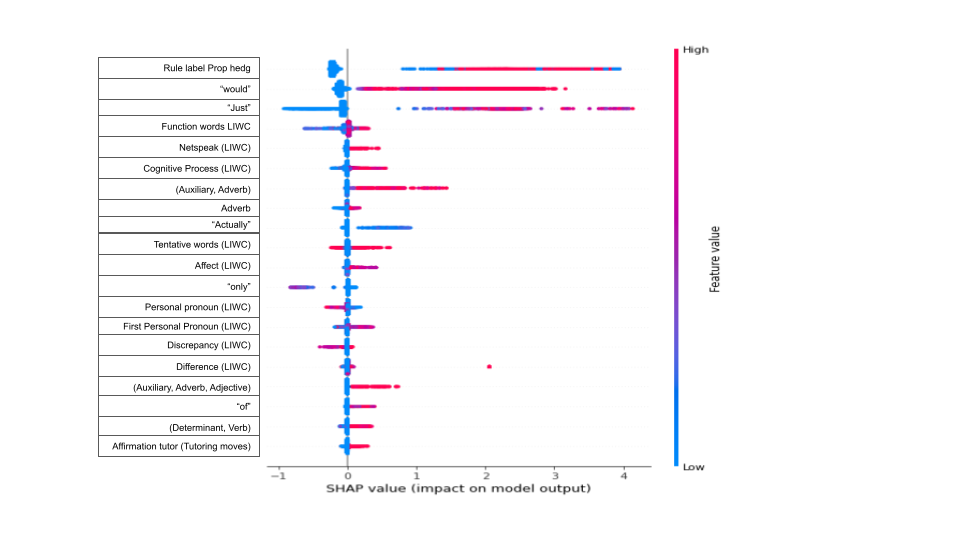}
    \caption{Averaged contribution of features to the detection of "Propositional hedges", as indicated by SHAP.}
    \label{fig:5}
\end{figure*}

\begin{figure*}[t]
    \centering
    \includegraphics[scale=0.5]{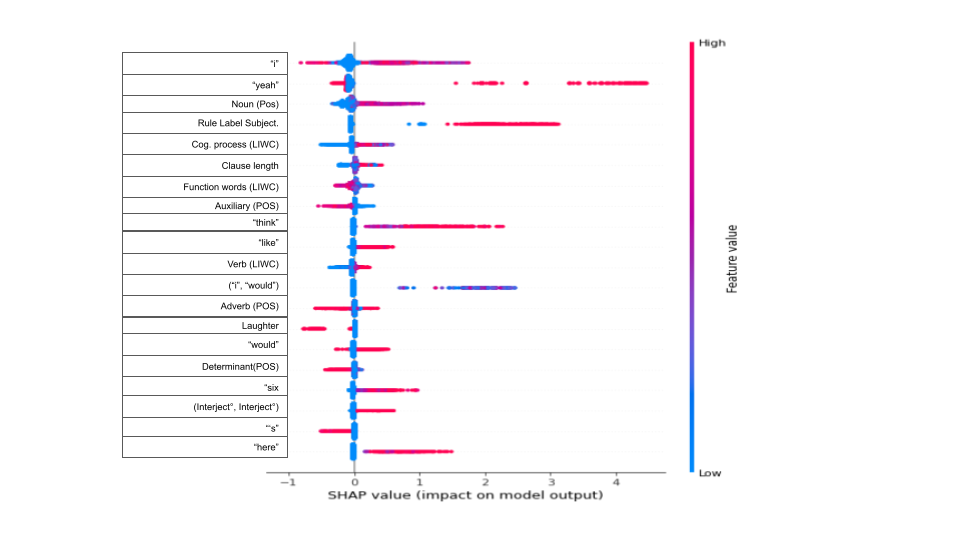}
    \caption{Averaged contribution of features to the detection of "Subjectivizers", as indicated by SHAP.}
    \label{fig:6}
\end{figure*}

\begin{table*}[t]
  \tabcolsep=0.05cm
  \tiny
  \centering
  \begin{tabular}{c|cccccccccccc} \toprule 
    Models & RB & MLP (KDF) & MLP (PTE) & MLP (K+P) & CNN (PTE) & LSTM (KDF) & LSTM(PTE) & LSTM (K+P) & BERT (PTE) & LGB (KDF) & LGB (PTE) & LGB (K+P) \\ \midrule 
    Rule-based & & No & Yes & No & No & No & Yes & No & No & Yes & Yes & No \\ 
    MLP (KDF) & No & & Yes & No & No & No & Yes & No & No & Yes & Yes & No \\ 
    MLP (PTE) & Yes & Yes & & Yes & Yes & Yes & No & Yes & Yes & Yes & No & Yes \\
    MLP (KDF + PTE) & No & No & Yes & & No & No & Yes & No & Yes & Yes & Yes & Yes \\
    Attention-CNN (PTE) & No & No & Yes & No & & No & Yes & No & Yes & Yes & Yes & Yes \\
    LSTM (KDF) & No & No & Yes & No & No & & Yes & No & Yes & Yes & Yes & Yes \\
    LSTM(PTE) & Yes & Yes & No & Yes & Yes & Yes & & Yes & Yes & Yes & Yes & Yes \\
    LSTM (KDF + PTE) & No & No & Yes & No & No & No & Yes & & Yes & Yes & Yes & Yes \\
    BERT (PTE) & No & No & Yes & Yes & Yes & Yes & Yes & Yes & & Yes & Yes & No \\
    LGBM (KDF) & Yes & Yes & Yes & Yes & Yes & Yes & Yes & Yes & Yes & & Yes & Yes \\
    LGBM (PTE) & Yes & Yes & No & Yes & Yes & Yes & Yes & Yes & Yes & Yes & & Yes \\
    LGBM (KDF + PTE) & No & No & Yes & Yes & Yes & Yes & Yes & Yes & No & Yes & Yes & \\ \bottomrule
  \end{tabular}
  \caption{Significance table for the 3-classes part of Table \ref{tab:4}. "Yes" means that the difference is statistically significant.}
  \label{tab:8}
\end{table*}
    
\begin{table*}[t]
  \tabcolsep=0.05cm
  \tiny
  \centering
  \begin{tabular}{c|cccccccccccc} \toprule 
    Models & RB & MLP (KDF) & MLP (PTE) & MLP (K+P) & CNN (PTE) & LSTM (KDF) & LSTM(PTE) & LSTM (K+P) & BERT (PTE) & LGB (KDF) & LGB (PTE) & LGB (K+P) \\ \midrule 
    Rule-based & & No & Yes & Yes & No & Yes & Yes & No & No & Yes & Yes & Yes \\ 
    MLP (KDF) & No & & Yes & Yes & No & Yes & Yes & Yes & No & Yes & Yes & No \\ 
    MLP (PTE) & Yes & Yes & & Yes & Yes & Yes & Yes & Yes & Yes & Yes & Yes & Yes \\
    MLP (KDF + PTE) & Yes & Yes & Yes & & No & No & Yes & No & Yes & Yes & Yes & Yes \\
    Attention-CNN (PTE) & No & No & Yes & No & & No & Yes & No & No & Yes & Yes & Yes \\
    LSTM (KDF) & Yes & Yes & Yes & No & No & & Yes & No & No & Yes & Yes & Yes \\
    LSTM(PTE) & Yes & Yes & Yes & Yes & Yes & Yes & & Yes & Yes & Yes & Yes & Yes \\
    LSTM (KDF + PTE) & Yes & Yes & Yes & No & No & No & Yes & & Yes & Yes & Yes & Yes \\
    BERT (PTE) & No & No & Yes & Yes & No & Yes & Yes & Yes & & Yes & Yes & No \\
    LGBM (KDF) & Yes & Yes & Yes & Yes & Yes & Yes & Yes & Yes & Yes & & Yes & Yes \\
    LGBM (PTE) & Yes & Yes & Yes & Yes & Yes & Yes & Yes & Yes & Yes & Yes & & Yes \\
    LGBM (KDF + PTE) & Yes & No & Yes & Yes & Yes & Yes & Yes & Yes & No & Yes & Yes & \\ \bottomrule
  \end{tabular}
  \caption{Significance table for the 4-classes part of Table \ref{tab:4}. "Yes" means that the difference is statistically significant.}
  \label{tab:9}
\end{table*}

\begin{table*}[t]
  \tabcolsep=0.2cm
  \small
  \centering
  \begin{tabular}{c|ccccccc} \toprule 
    Models & Label RB & + 1-gram and 2-gram & + POS & + LIWC & + TM  & + Nonverbal \\ \midrule 
    Label RB & & Yes & Yes & Yes & Yes & Yes\\ 
    + 1-gram and 2-gram & Yes & & No & No & No & No \\ 
    + POS & Yes & No & & No & No & No\\
    + LIWC & Yes & No & No & & No & No \\
    + TM & Yes & No & No & No & & No \\
    + Nonverbal & Yes & No & No & No & No & \\ \bottomrule
  \end{tabular}
  \caption{Significance table for the 3-classes part of Table \ref{tab:5}. "Yes" means that the difference is statistically significant.}
  \label{tab:10}
\end{table*}

\begin{table*}[t]
  \tabcolsep=0.2cm
  \small
  \centering
  \begin{tabular}{c|ccccccc} \toprule 
    Models & Label RB & + 1-gram and 2-gram & + POS & + LIWC & + TM  & + Nonverbal \\ \midrule 
    Label RB & & Yes & Yes & Yes & Yes & Yes\\ 
    + 1-gram and 2-gram & Yes & & No & No & No & No \\ 
    + POS & Yes & No & & No & No & No\\
    + LIWC & Yes & No & No & & No & No \\
    + TM & Yes & No & No & No & & No \\
    + Nonverbal & Yes & No & No & No & No & \\ \bottomrule
  \end{tabular}
  \caption{Significance table for the 4-classes part of Table \ref{tab:5}. "Yes" means that the difference is statistically significant.}
  \label{tab:11}
\end{table*}

\end{document}